%% file: main.tex
\documentclass[10pt,twocolumn,letterpaper]{article}

\usepackage[pagenumbers]{cvpr} %

\input{preamble}

\definecolor{cvprblue}{rgb}{0.21,0.49,0.74}
\usepackage[pagebackref,breaklinks,colorlinks,allcolors=cvprblue]{hyperref}

\def\paperID{*****} %
\def\confName{CVPR}
\def\confYear{2026}

\title{Deepfake-Eval-2024: A Multi-Modal In-the-Wild Benchmark of Deepfakes Circulated in 2024}

\author{
  Nuria Alina Chandra$^{1}$ \quad
  Hannah Lee$^{1}$ \quad
  Ryan Murtfeldt$^{1,2}$ \quad
  Lin Qiu$^{1,2}$ \quad
  Arnab Karmakar$^{1,2}$ \\
  Emmanuel Tanumihardja$^{1,2}$ \quad
  Kevin Farhat$^{3}$ \quad
  Ben Caffee$^{1,2}$ \quad
  Changyeon Lee$^{4}$ \quad
  Jongwook Choi$^{5}$ \\
  Sejin Paik$^{6}$ \quad
  Aerin Kim$^{1,7}$ \quad
  Oren Etzioni$^{1,2}$ \\[2ex]
  $^{1}$TrueMedia.org \quad
  $^{2}$University of Washington \quad
  $^{3}$Allen Institute for Artificial Intelligence \\
  $^{4}$University of Maryland \quad
  $^{5}$Chung-Ang University \quad
  $^{6}$Georgetown University \quad
  $^{7}$Miraflow AI
}

\begin{document}
\maketitle

\input{sections_cvpr/0_abstract}
\input{sections_cvpr/1_introduction}

\input{sections_cvpr/2_related_work}
\input{sections_cvpr/3_dataset}

\input{sections_cvpr/4_experiments}

\input{sections_cvpr/5_errors}
\input{sections_cvpr/6_conclusion}

\input{sections_cvpr/8_acknowledgements}

{
    \small
    \bibliographystyle{ieeenat_fullname}
    \bibliography{main}
}

\input{sections_cvpr/9_supp_materials}

\end{document}

%% file: preamble.tex
\newcommand{\TODO}[1]{\textbf{\color{red}[TODO: #1]}}
\renewcommand{\TODO}[1]{}

\newcommand{\TrueMedia}{TrueMedia}

\usepackage{pifont}

\usepackage{placeins}
\usepackage{longtable}

\usepackage{array}
\newcolumntype{z}[1]{>{\centering\arraybackslash\hspace{0pt}}m{#1}}

\newcommand{\tabletext}[1]{%
    {\small \\ #1\par}
}

\usepackage[accsupp]{axessibility}  %

%% file: sections_cvpr/0_abstract.tex
\begin{abstract}
In the age of increasingly realistic generative AI, robust deepfake detection is essential for mitigating fraud and disinformation. While many deepfake detectors report high accuracy on academic datasets, we show that these academic benchmarks are out of date and not representative of real-world deepfakes. We introduce Deepfake-Eval-2024, a new deepfake detection benchmark consisting of in-the-wild deepfakes collected from social media and deepfake detection platform users in 2024. Deepfake-Eval-2024 consists of 45 hours of videos, 56.5 hours of audio, and 1,975 images, encompassing the latest manipulation technologies. The benchmark contains diverse media content from 88 different websites in 52 different languages. We find that the performance of open-source state-of-the-art deepfake detection models drops precipitously when evaluated on Deepfake-Eval-2024, with AUC decreasing by 50\% for video, 48\% for audio, and 45\% for image models compared to previous benchmarks. We also evaluate commercial deepfake detection models and models finetuned on Deepfake-Eval-2024, and find that they have superior performance to off-the-shelf open-source models, but do not yet reach the accuracy of deepfake forensic analysts. %
\end{abstract}

%% file: sections_cvpr/1_introduction.tex
\begin{figure*}[ht]
    \centering
    \includegraphics[width=0.95\textwidth]{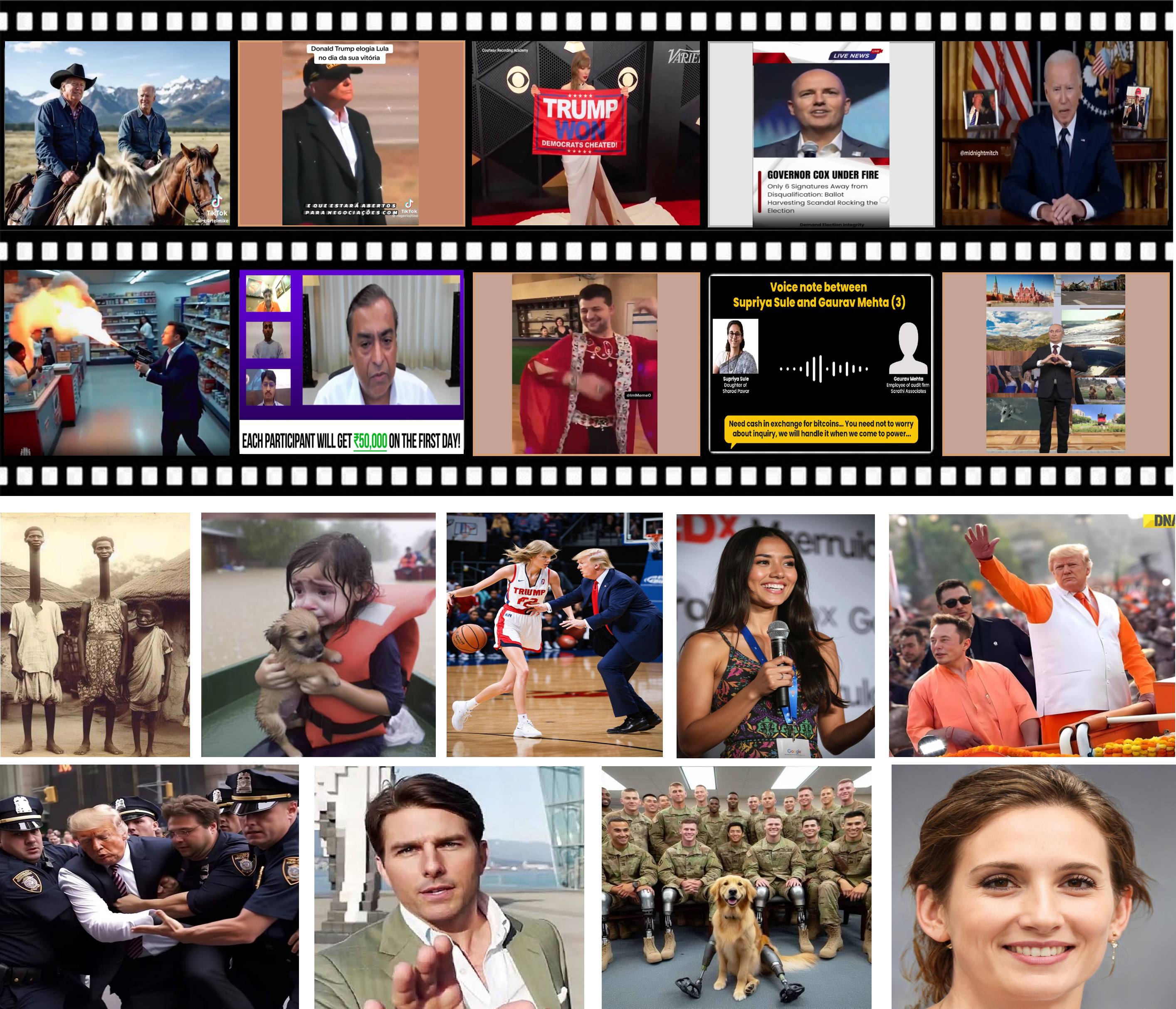} %
    \caption{Examples of Deepfake-Eval-2024 video and audio (rows 1–2), and images (rows 3–4), demonstrating a diversity of content styles and generation techniques, including lipsync, faceswap, and diffusion. Images have been resized for presentation.}
    \label{fig:example} %
\end{figure*}

\section{Introduction}
Advances in generative AI models have precipitated a surge of highly realistic deepfakes, which have been used to fabricate messages from politicians \cite{suwajanakorn2017synthesizing}, create non-consensual pornographic content \cite{umbach2024nonconsensualsyntheticintimateimagery}, spread misinformation \cite{helmus22disinformation}, and damage reputations \cite{reputation_tv}, harming lives, businesses, and nations \cite{dhs:deepfakes-threats}. Between 2023 and 2024, there was a fourfold increase in the number of deepfakes detected in fraud \cite{sumsub:2024}, and in 2023 alone, an estimated 500,000 deepfakes were shared on social media websites \cite{sumsub:2024}.

Recent research has shown that people are no longer able to determine whether media is AI-generated or real \cite{frank2023representativestudyhumandetection}. Thus, the development of accurate and automated deepfake detection methods has become essential for mitigating the harmful effects of deepfakes. Many deepfake detection models have already been developed, such as GenConViT \cite{wodajo2023genconvit} for video, AASIST \cite{Jung2021AASIST} for audio, and NPR \cite{tan2023rethinking_npr} for image deepfakes, all of which perform extremely well on the academic datasets that they were originally benchmarked on, with AUC values approaching one (Table \ref{tab:results-opensource}). However, these datasets are not representative of deepfakes circulating on social media, because many use outdated manipulation techniques (\eg, FaceForensics++ \cite{faceforensicspp} and ForgeryNet \cite{he2021forgerynet}) with human differentiable fakes (\eg, faces not centered on heads in ForgeryNet). Further, most existing synthetic datasets and in-the-wild datasets have limited content diversity (\eg, exclusively single-scene videos containing a limited number of body poses \cite{barrington2024deepspeakdatasetv10, DFDC2020, jiang2020deeperforensics1}, or exclusively English audio \cite{ khalid2021fakeavceleb, Liu_2023_asvspoof21, muller2022does_in-the-wild, jung2024spoofceleb}). In-the-wild deepfake are also limited, with the most recent prior datasets published in 2022 (audio) \cite{muller2022does_in-the-wild}, and 2023 (video) \cite{cho2023towards}, prior to widespread adoption of techniques such as stable diffusion \cite{rombach2022_stable_diffusion_v1} and the release of commercial models such as ElevenLabs voice conversion \cite{elevenlabs2025}.

To address the limitations of existing deepfake detection benchmarks, we present Deepfake-Eval-2024, a dataset collected from social media and the free deepfake detection platform \TrueMedia.org. Each item in Deepfake-Eval-2024 comes from a social media or \TrueMedia.org user who flagged the media as potentially AI-manipulated in 2024. As a result, Deepfake-Eval-2024 is smaller than synthetic datasets, but much more diverse and directly representative of deepfakes in the wild. We summarize our contributions with the following:
\begin{itemize}[noitemsep,topsep=0pt]
    \item We release a challenging multimodal in-the-wild deepfake detection benchmark. 
     \item To our knowledge, this is the first in-the-wild dataset that includes video, audio, and images, and it is the largest and most diverse in-the-wild deepfake detection dataset. 
     \item We evaluate state-of-the art deepfake detectors (both open-source and commercial) on contemporary in-the-wild data demonstrating their limitations and suggesting directions for future work.
 \end{itemize}

%% file: sections_cvpr/2_related_work.tex
\section{Related Work}
Deepfake datasets have not kept up with the fast-moving field of AI content generation. This is particularly true of in-the-wild datasets; prior to the release of Deepfake-Eval-2024, the only other in-the-wild audio deepfake dataset was published in 2022 \cite{muller2022does_in-the-wild}, with fewer hours of audio than Deepfake-Eval-2024. To the best of our knowledge, prior to Deepfake-Eval-2024, there were no other in-the-wild image-focused deepfake datasets. The most recent in-the-wild video datasets were from 2021 \cite{pu2021deepfake_dfw}, and 2023 \cite{cho2023towards}. Supplementary Tables \ref{tab:related-work_video}, \ref{tab:related-work_audio}, \ref{tab:related-work_image} provide a detailed survey of popular datasets that were released prior to Deepfake-Eval-2024. 

Past in-the-wild datasets collect deepfakes by searching for media explicitly labeled as such (\eg, \cite{cho2023towards} queried `deepfake' in four languages), yet most real-world deepfakes posing security risks are unlabeled \cite{helmus2022artificial}. Deepfake-Eval-2024 instead draws from media flagged as potentially AI-manipulated by social media and \TrueMedia.org users (primarily journalists), making it more representative of security-relevant deepfakes. It also vastly expands diversity: prior in-the-wild datasets used at most four sources \cite{cho2023towards}, while Deepfake-Eval-2024 spans 88 web sources and 42 languages in our audio dataset and 52 different languages in combined video and audio datasets (Figure \ref{fig:all-origins-language}), compared to a maximum of two languages in prior audio datasets.

Most existing deepfake datasets are synthetically generated, enabling large scale (\eg, ForgeryNet \cite{he2021forgerynet}, DFDC \cite{DFDC2020}, and AV-Deepfake1M \cite{cai2024av-deepfake1m}), but they fail to capture the distribution of deepfakes circulating on social media. Synthetic video datasets apply a limited set of manipulation techniques to curated, often highly structured real videos \cite{barrington2024deepspeakdatasetv10, DFDC2020, jiang2020deeperforensics1}, focus exclusively on face manipulations, and image datasets largely repurpose video frames \cite{faceforensicspp, zi2020wilddeepfake} or target general AI-generated content rather than deepfakes \cite{Bird2023CIFAKEIC, wang2023dire_diffusionforensics}. Deepfake-Eval-2024 addresses these gaps with diverse real-world content including varied settings, actions, and manipulations beyond the face (Figure \ref{fig:example}).

%% file: sections_cvpr/3_dataset.tex
\section{Dataset}\label{sec:dataset}
\begin{table}
  \caption{Deepfake-Eval-2024 Summary Statistics}
  \label{tab:summary_stats}
  \makebox[\linewidth]{
      \centering
      \begin{tabular}{lccc}
        \toprule
        \textbf{Modality} & \textbf{Size} & \textbf{Avg Resolution}  \\
        \midrule
        Video   & 45.1 hrs & 30.05 FPS,  576$\times$720 px \\
        Audio   & 56.5 hrs & 44.66 kHz\\
        Image   & 1,975 images  & 1,024\ensuremath{\times}1,024 px \\
        \bottomrule
      \end{tabular}
  }
\end{table}

Deepfake-Eval-2024 is composed of 45 hours of videos and 56.5 hours of audio and 1,975 images. (Table \ref{tab:summary_stats} contains abbreviated summary statistics, with complete statistics stratified by label available in Supplementary Tables \ref{append-tab:video_stats}, \ref{append-tab:audio_stats}, and \ref{append-tab:image_stats}.) The data includes real, AI-generated, and AI-manipulated content. Audio data includes audio from videos, in addition to audio-only media. The majority of video data has corresponding labeled audio.

An ideal benchmark for deepfake detection is representative of the real-world threat of deepfakes. This requires the following criteria: 1) it contains fake and real content that is difficult for humans to categorize, 2) it includes all popular generative techniques used for deepfake generation, and 3) it has diverse content, representative of media shared on the internet. We will show that Deepfake-Eval-2024 meets all three criteria through its unique data collection approach. All data was collected through the deepfake detection platform \TrueMedia.org and social media content moderation forums.

\subsection{Data Collection}

\begin{figure*}[htbp]
    \centering
    \begin{minipage}{0.45\textwidth}
        \centering
        \includegraphics[width=\textwidth]{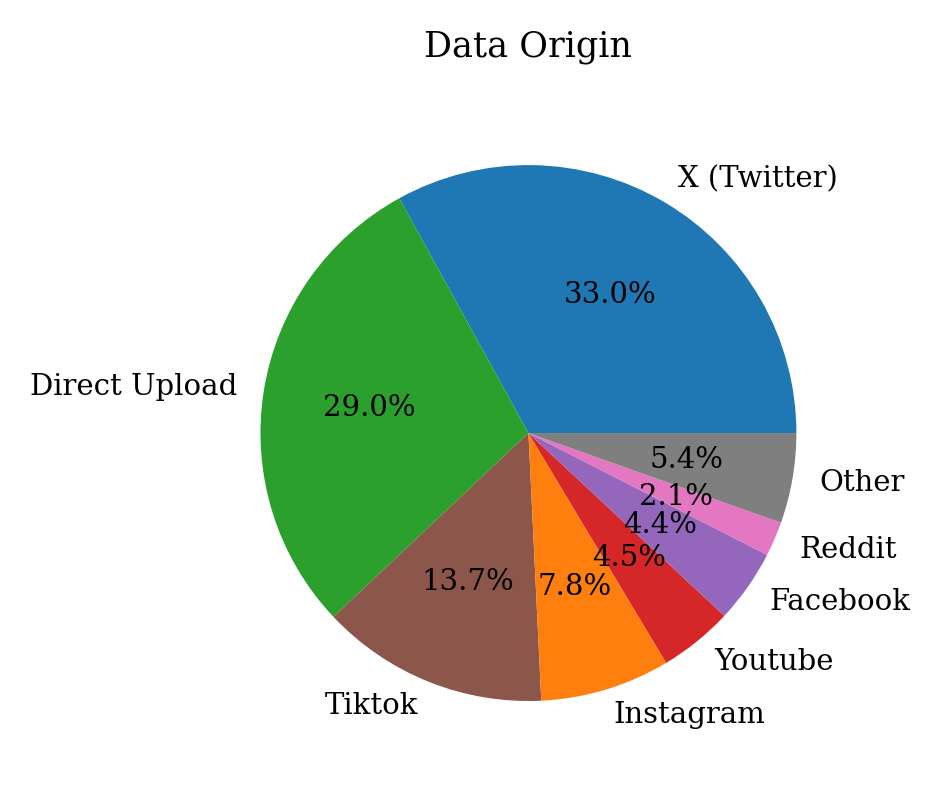}
    \end{minipage}
    \hfill
    \begin{minipage}{0.45\textwidth}
        \centering
        \includegraphics[width=\textwidth]{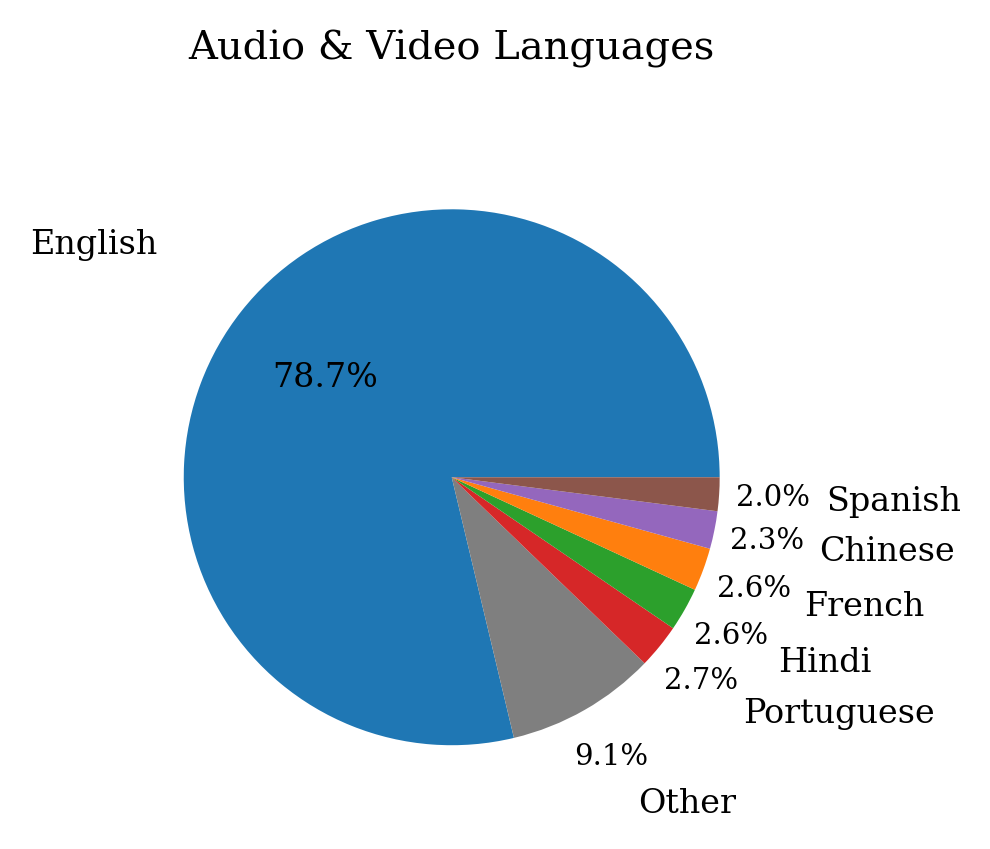}
    \end{minipage}
    \hfill
    \caption{Distribution of data origins and languages in Deepfake-Eval-2024. In total, media was shared from 88 different web-domain names. The dataset also contains a total of 52 different languages (42 languages in Deepfake-Eval-2024-audio and 49 languages in Deepfake-Eval-2024-video). Languages were identified using speech recognition model Whisper \cite{radford2022robustspeechrecognitionlargescale}.}
    \label{fig:all-origins-language}
\end{figure*}

\textbf{Data Sources}. The deepfake detection platform \TrueMedia.org was a non-profit application used by journalists and fact-checkers starting in April 2024, and used by the general public starting in September 2024. Deepfake-Eval-2024 includes data uploaded to \TrueMedia.org. Users provided a social media link or directly uploaded content to be checked for AI-manipulation. We also created a bot on X (previously known as Twitter) that allowed users to add content to our platform by tagging the bot. The code for this bot is available at \texttt{https://github.com/truemediaorg/socialbot}. In addition, we uploaded X posts that had been flagged by X Community Notes as potentially manipulated media. The top five most common data sources of Deepfake-Eval-2024 are X, direct upload, TikTok, Instagram, and Youtube (Figure \ref{fig:all-origins-language}, Supp. Figure \ref{ext-fig:all-origins-separate}).

\textbf{Collection Ethics}. \TrueMedia.org users were informed in the Terms of Use that `by sharing your content, you agree to follow our content terms and you agree everything you share is consistent with these terms. Anything you share will not be private and can be used by us, our partners, and others we work with, in lots of different ways.' To the best of our knowledge the dataset does not contain any personal identification information (i.e. social security numbers, emails, financial information, or medical records), and due to the nature of our data (non-textual), it is highly unlikely that we have incidentally collected such information through either \TrueMedia.org or other web-sources. 

\textbf{Data Attributes}. Our data collection method ensures that Deepfake-Eval-2024 is a \textbf{challenging} dataset. Users often brought media to \TrueMedia.org when it could not be easily identified as real or fake by a human. Thus, we estimate that Deepfake-Eval-2024 has a greater proportion of challenging examples in both real and fake categories than prior datasets. 

\textbf{Dataset Diversity}. Collecting data through social media and deepfake detection platform users also provides increased diversity with respect to generative models, ethnicity, language, and content. Deepfake-Eval-2024 is a sample of currently circulating AI-generated content. Thus, we estimate that our dataset includes AI-generated and manipulated content from every type of commonly used contemporary deepfake generation model. Further, \TrueMedia.org users came from all over the world, resulting in increased ethnic and linguistic diversity in our dataset. Our dataset is 78.7\% English and includes a total of 52 different languages (Figure \ref{fig:all-origins-language}). We estimate racial demographics by labeling 10\% of the visual data for perceived race/ethnicity. Visual data is estimated to be 54.6\% white and 45.4\% non-white with full results in Supp. Table \ref{app:demographic-estimates}. The content of the media itself has larger variations than the standardized content typically found in academic datasets. The diversity in data origins (Figure \ref{fig:all-origins-language}, Supp. Figure \ref{ext-fig:all-origins-separate}) results in a wide variety of different media styles, including videos of political speeches and self-shot content-creators, images of large crowds and close-up portraits, and audio clips of debating politicians. 

\subsection{Data Filtering} We remove duplicate data using a combination of manual review and hash functions. We include cases where two pieces of media have minor variations and thus appear to be the same (\eg, different cropping of the same video). In order to tailor our datasets to evaluate deepfake detection models, we remove images and videos that do not contain photorealistic faces. This resulted in the removal of cartoons, art, and scenes without humans. We use GPT-4o (version 2024-08-06) to identify images with photorealistic faces. We note that GPT-4o's responses have high precision but also a high false negative rate. To account for this, we manually review all images marked as non-photorealistic faces by GPT-4o. To identify videos with photorealistic faces we first use the dlib face detection library \cite{dlib} to determine whether each frame contains a face or not. Videos where no faces are detected are then reviewed manually to check for missed faces.

\subsection{Data Labeling}\label{subsec:data-labeling}

\begin{table*}
  \caption{Inter-rater disagreement statistics}
  \label{tab:labeling-disagreement}
  \makebox[\linewidth]{
  \centering
  \begin{tabular}{lccccc}
    \toprule
    \textbf{Modality} & \textbf{N Checked} & \textbf{Total Disagreement} & \textbf{Real vs Fake} & \textbf{Real vs Unknown} & \textbf{Fake vs Unknown} \\
    \midrule
    Video & 243 & 6.6\% & 2.1\% & 3.7\% & 0.8\% \\
    Audio & 342 & 7.9\% & 0.6\% & 3.5\% & 3.8\% \\
    Images & 269 & 9\% & 0\% & 3\% & 6\% \\
    \bottomrule
  \end{tabular}
  }
\end{table*}

\textbf{Label Definition}. We label media as fake if it was AI-generated or manipulated. We choose to define our labels this way despite the challenge of differentiating AI-manipulated content from traditional forms of manipulation so that this dataset can be used to benchmark deepfake detection models which are trained to differentiate between AI-generated and real content.  

\textbf{Labeling Methodology}. The labeling team consisted of seven people conducting deepfake forensic analysis: three experienced AI-generated content labelers and four machine learning research interns. The team met regularly to discuss the taxonomy, verification process, and edge cases. Our verification process consisted of locating original sources using reverse image search or searching the web using quotes or situation descriptions, then confirming the trustworthiness of the source, and scanning media for characteristics of AI-generated media (Appendix \ref{app:labeling-criteria}). See Appendix \ref{app:verification-process} for our detailed verification process. The team labeled each piece of media as ``fake,'' ``real,'' or ``unknown'' when the appropriate label was not clearly discernible from authenticated sources or media characteristics. Media labeled ``unknown'' were excluded.

\textbf{Deepfake Forensic Analysis}. We rely on articles published by professional fact-checking organizations such as Snopes \cite{Snopes2025} and AFP Fact Check \cite{AFPFactCheck2025} to confirm if AI manipulation was used. Additionally, we utilize community moderation platforms like X Community Notes to locate and review primary sources. We conduct source context verification by comparing social media posts with original materials such as full-length videos to verify if media has been manipulated. We further scrutinize the media for evidence of AI manipulation using specific forensic markers. For example, we rely on the synchronization of mouth movements with vocal sounds as a primary measure of authenticity in video and audio media. For videos and images, face-swaps are classified as fake if they were created after 2023, and unknown otherwise. This time separation was chosen based on research that suggests that the majority of face-swaps created in 2023 or later use AI \cite{iproov2024threat}. Other common indicators of AI manipulation, including anatomical implausibilities, sociocultural implausibilities, and stylistic artifacts, are identified based on the framework created by Kamali et al. \cite{kamali2024how}. When the label of audio media cannot be determined from sources, due to the challenge of differentiating AI generated audio from non-AI generated voice impersonators, media is marked as fake if and only if there are both audible traits indicating that it is fake (\eg sociocultural implausibilities), in addition to at least two commercial audio detectors predicting the media as fake. Audio detectors alone are never used to label. We note that this labeling approach results in audio labels that are correlated with existing detectors, which is a limitation of the audio dataset. However, this approach is necessary, as it is impossible to differentiate between socioculturally-implausible audio made by highly skilled impersonators (real audio) and AI-generated audio (see Jordan Peele's highly accurate Barack Obama impersonation which uses real audio and AI-manipulated video as an example \cite{peele2018deepfake, romano2018peele}). See Appendix \ref{app:labeling-criteria} for complete labeling taxonomy and media examples.

\textbf{Labeling Accuracy}. To assess the consistency and quality of our annotations, the labeling team lead double-reviewed a random sample of 10\% of the data. Annotations created by the team lead were excluded to avoid self-assessment bias. For videos, we find a 6.6\% disagreement between labelers, with the largest discrepancy between real and unknown at 3.7\%. For audio, we find a 7.9\% disagreement between labelers, with the largest discrepancies between real and unknown at 3.5\% and between fake and unknown at 3.8\%. And for images, we find a 9\% disagreement between labelers, with the largest discrepancy between fake and unknown at 6\%. (See Table \ref{tab:labeling-disagreement} for complete disagreement breakdown.) This disagreement between trained labelers represents the challenge of the task. Given that the inter-labeler disagreement was consistently below 10\% , we posit that deepfake detection models should be able to achieve at least 90\% accuracy on Deepfake-Eval-2024, and likely higher given that real vs. fake disagreement is always below 2.5\%. 

Common labeling errors include: differentiating between dubbed videos (where the video has not been AI-manipulated, and thus is real), and lipsynced videos (where the video has been AI-manipulated to make the mouth match new words and thus should be marked as fake); determining if audio sound is synthetic; and missing anatomical or sociocultural implausibilities.

%% file: sections_cvpr/4_experiments.tex
\section{Experiments}\label{sec:4-experiments}

\textbf{Model Selection}. To evaluate the state-of-the-art of deepfake detection on real world in-the-wild deepfakes, we test an array of open-source deepfake detection models on Deepfake-Eval-2024. We select standard models that encompass the primary deepfake detection model architectures associated with each modality. Models were also selected based on the availability of pretrained model weights and runnable training code. All models were chosen prior to experimentation, and no models were omitted on the basis of performance.  

\textbf{Open-Source Models}.
For each modality, we evaluate three different open-source deepfake detection models on the modality-appropriate Deepfake-Eval-2024 data. For image detection we include a single layer perceptron with a CLIP \cite{radford2021learning-clip} backbone (UFD \cite{ojha2023fakedetect_ufd}), a model based on diffusion inversion (DistilDIRE \cite{lim2024distildire}), and a convolutional neural network (NPR \cite{tan2023rethinking_npr}). For audio detection we include a spectro-temporal graph attention network (AASIST \cite{Jung2021AASIST}), a convolutional neural network applied to raw waveforms (RawNet2 \cite{tak2021endtoendantispoofingrawnet2}), and a model with a self-supervised component (P3 from Wang et al.~\cite{wang2023largescalevocodedspoofeddata_model-id-p3}). We choose video models that have a generative convolutional vision transformer (GenConViT \cite{wodajo2023genconvit}), a temporal convolutional network (FTCN \cite{zheng2021ftcn}), and a model that evaluates style latent vectors (Styleflow \cite{choi2024styleflow}). We also evaluated all open source multimodal models with runable code that we could find, AVF \cite{feng2023self} and FGI \cite{astrid2024detecting}. We use the code and preprocessing approaches described in the original publications. To adapt to open-source audio models with a limit of four seconds, we split Deepfake-Eval-2024 audio files into four-second segments and report performance on these segments.

\textbf{Commercial Models}. We evaluate commercially available deepfake detection models from companies that partnered with \TrueMedia.org: Hive, Reality Defender, Pindrop, AI or Not, Hiya, Fraunhofer, and Sensity AI. Many companies provide multiple models. In total, we evaluate 22 different commercial models (six video, eight audio, and eight image models). We anonymize the performance of the models and providers to comply with contractual agreements. Commercial models are evaluated using their latest available versions as of December 2024. All vendors were blind to the test data. Due to the high per-query cost of commercial vendors, we were unable to evaluate all commercial models on the entirety of Deepfake-Eval-2024. Instead, we evaluate all models on a subset of Deepfake-Eval-2024, and then evaluate the top three models for each modality on the entire Deepfake-Eval-2024. We report the performance of the top commercial model for each modality in Table \ref{tab:results-commercial}, and Supp. Table \ref{app-table:commercial-results}.

\textbf{Evaluation Metrics}. Past deepfake detection publications have selectively focused on specific metrics (\eg AUC in \cite{yan2024df40}, accuracy in \cite{ojha2023fakedetect_ufd}, and EER in \cite{wang2024asvspoof5}). Selective reporting on specific metrics makes comparison across publications challenging and does not offer a comprehensive view of model performance. As such, we evaluate the performance of models on Deepfake-Eval-2024 through a wide variety of metrics focusing on AUC, F1-score, and accuracy, with full metrics including precision, recall, false positive rate, and false negative rate available in the supplementary materials. We also report EER for audio models in the supplementary, as it is common in audio deepfake literature. Some open-source and commercial models fail to run on all media files due to model constraints (\eg, media length limits, or requirements for a face to be detected in a certain number of frames). When a model fails to produce a prediction, we exclude this file when calculating the metrics for the associated model. 

\textbf{Evaluation on Previous Benchmarks}. We compare the performance of open-source models on our dataset to the performance of each model on the test datasets reported in its original publication (Table \ref{tab:results-opensource}). To account for different reporting metrics used across publications, we recompute predictions on the originally published test datasets to provide a full array of evaluation metrics. Where multiple test datasets were reported in the original publication, we compute results on as many of the datasets as possible and report average metrics across these test datasets.

\textbf{Finetuning}. To determine if real-world deepfake detection performance can be improved by training on representative data, we finetune all open-source models on 60\% of Deepfake-Eval-2024, and evaluate the performance on the remaining 40\% of the data (Table \ref{tab:finetune-combined}). This split mirrors real-world scenarios where models must generalize from limited training data to detect unseen deepfake techniques. We finetune each model following the original authors' recommended training procedures and hyperparameters where available, using early stopping to avoid overfitting.

\begin{table*}[htbp]
    \input{tables/opensource-tables-combined}
    \label{tab:results-opensource}
    \\[0.5em]
    \small{Original publication test data includes the following datasets for each model. Where multiple datasets are specified, the reported metrics are averages over these datasets. Evaluations were re-run on original publication test data if metrics were not provided in the original publication. GenConViT: \cite{faceforensicspp}, \cite{DFDC2020}, \cite{celeb_df}, \cite{korshunov2018_DeepfakeTIMIT}; FTCN: \cite{celeb_df}; Styleflow: \cite{celeb_df}, \cite{DFD_dufour2019}, \cite{faceshifter}, \cite{deeperforensics}; AASIST, RawNet2, and P3 were all evaluated on the LA eval set of ASVspoof2019 \cite{wang2020asvspoof2019}; UFD: \cite{wang2019cnngenerated_forensynth} and subsets of LAION-400M \cite{schuhmann2021laion400mopendatasetclipfiltered} and AI generated images from latent diffusion models \cite{rombach2022_stable_diffusion_v1}, Glide \cite{nichol2022glidephotorealisticimagegeneration}, and DALL·E mini \cite{Dayma_DALL·E_Mini_2021} provided by the original publication \cite{ojha2023fakedetect_ufd}; DistilDIRE: ImageNet and AI generated images from Stable Diffusion v1 \cite{rombach2022_stable_diffusion_v1} and ADM \cite{dhariwal2021diffusion_ADM} as specified in the original publication \cite{lim2024distildire}; NPR: \cite{chuangchuangtan-GANGen-Detection}, \cite{wang2023dire_diffusionforensics}, and the dataset from \cite{ojha2023fakedetect_ufd}. Complete performance metrics on Deepfake-Eval-2024 are available in Supp. Table \ref{append-tab:off-shelf-combined}.}
\end{table*}

\subsection{Open-Source Model Performance}

\textbf{All off-the-shelf open-source models perform poorly on Deepfake-Eval-2024}. The maximum AUC of open-source models across modalities and models was 0.58  (Table \ref{tab:results-opensource}, Supp. Table \ref{append-tab:off-shelf-combined}). The performance of multimodal models is similarly poor (Supp. Table 
\ref{append-tab:multimodal-results}). Further, many off-the-shelf models have an AUC close to 0.5, the same as random guessing, suggesting that these models perhaps learned to predict deepfakes based on correlations that were present in academic training datasets but do not exist in contemporary real-world data.

\textbf{Performance on Deepfake-Eval-2024 is considerably lower than on previous benchmarks}. The poor performance of open-source models on Deepfake-Eval-2024 offers a stark contrast to the exceptional performance of these models on the datasets that they were originally tested on (right side of Table \ref{tab:results-opensource}). We observe an average drop in AUC of 50\%
for video, 48\% for audio, and 45\% for image models when evaluated on Deepfake-Eval-2024, as compared to the academic datasets that the models were originally tested on. This drastic difference in performance suggests that the academic deepfake detection datasets which the models were trained to perform well are not representative of the threat of contemporary deepfakes, underscoring the importance of up-to-date, challenging, in-the-wild deepfake datasets like Deepfake-Eval-2024. 

\subsection{Finetuned Model Performance}
\begin{table}[htbp]

\input{tables/finetune-result-combined}
        \label{tab:finetune-combined}
\end{table}

\textbf{Models improve when finetuned on a subset of Deepfake-Eval-2024}. AUC for open-source models improves by an average of 57.6\% for video, 80.6\% for audio and 4.5\% for images (Table \ref{tab:finetune-combined}, Supp. Table \ref{append-tab:finetune-combined}). However, the degree of improvement varies across models, suggesting that some model architectures may be less suited to adapt to the challenges of real-world deepfake detection. For example, the simple single-layer UFD model learns to predict all data as fake after finetuning, and video model Styleflow also shows limited improvement in AUC. The limited improvement in image models is likely attributable to the relatively small amount of image finetuning data, which was insufficient to shift the highly paramaterized models of NPR and DistilDIRE towards the distribution of in-the-wild deepfakes. Although performance improves after finetuning, there is still significant room for further improvement, with the peak accuracy reaching 0.75 for videos, 0.86 for audio, and 0.63 for images, which is below the 90\% lower bound estimate of human deepfake analyst accuracy (Section \ref{subsec:data-labeling}). These results suggest that in addition to more representative training datasets, new model paradigms may be needed for robust and reliable deepfake detection.

\subsection{Commercial Model Performance}

\begin{table}[htbp]
    \input{tables/commercial-results}
    \label{tab:results-commercial}
\end{table}

\textbf{Top commercial models exceed the performance of open-source models}. Top commercial models considerably outperform off-the-shelf open-source models and finetuned image models, and perform slightly better than finetuned audio and video models. Commercial audio model performance is particularly strong, which is likely attributable to the limitations of the audio labeling approach as described in Section \ref{subsec:data-labeling}. No commercial models that we evaluated had an accuracy of 90\% or above, suggesting that commercial models still need improvement to reach the accuracy of human deepfake forensic analysts. In addition, we note that open-source models finetuned on a subset of Deepfake-Eval-2024 approach the accuracy of commercial models (finetuned video model GenConViT has an accuracy of 75\%, and finetuned audio model P3 from \cite{wang2023largescalevocodedspoofeddata_model-id-p3} has an accuracy of 86\%). This suggests that the competitive advantage of these commercial models may be derived primarily from training dataset curation.

%% file: tables/opensource-tables-combined.tex
\caption{Open-source model performance on Deepfake-Eval-2024 and original benchmarks}
\makebox[\linewidth]{
\centering
\begin{tabular}{c|c|cccc|cccc}
\toprule
& & \multicolumn{4}{c|}{Deepfake-Eval-2024} & \multicolumn{4}{c}{Original Publication Test Data} \\ [3pt]
\textbf{Modality} & \textbf{Model} & \textbf{AUC} & \textbf{Prec.} & \textbf{Recall} & \textbf{F1} & \textbf{AUC} & \textbf{Prec.} & \textbf{Recall} & \textbf{F1} \\
\midrule
Video & GenConViT \cite{wodajo2023genconvit} & 0.63 & 0.60 & 0.50 & 0.54 & 0.96 & 0.93 & 0.99 & 0.96 \\
& FTCN \cite{zheng2021ftcn}& 0.50 & 0.51 & 0.67 & 0.41 & 0.87 & 0.91 & 1.00 & 0.95 \\
& Styleflow \cite{choi2024styleflow}& 0.51 & 0.54 & 0.43 & 0.48 & 0.95 & 0.96 & 0.89 & 0.77 \\
\midrule
Audio & AASIST \cite{Jung2021AASIST} & 0.43 & 0.31 & 0.51 & 0.39 & 1.00 & 1.00 & 0.95 & 0.97 \\
& RawNet2 \cite{tak2021endtoendantispoofingrawnet2} & 0.53 & 0.66 & 0.39 & 0.49 & 0.99 & 0.60 & 0.99 & 0.74 \\
& P3 \cite{wang2023largescalevocodedspoofeddata_model-id-p3} & 0.58 & 0.36 & 1.00 & 0.53 & 1.00 & 1.00 & 0.96 & 0.98 \\
\midrule
Image & UFD \cite{ojha2023fakedetect_ufd} & 0.56 & 0.63 & 0.999 & 0.77 & 0.94 & 0.95 & 0.67 & 0.75 \\
& DistilDIRE \cite{lim2024distildire} & 0.52 & 0.64 & 0.87 & 0.74 & 0.99 & 0.99 & 0.98 & 0.98 \\
& NPR \cite{tan2023rethinking_npr} & 0.53 & 0.69 & 0.29 & 0.41 & 0.98 & 0.95 & 0.94 & 0.94 \\
\bottomrule
\end{tabular}
}

%% file: tables/finetune-result-combined.tex
\caption{Open-Source Model Finetuning Results. Complete performance metrics are available in Supp. Table \ref{append-tab:finetune-combined}.}
\makebox[\linewidth]{
\centering
\begin{tabular}{llcccc}
\toprule
\textbf{Modality} & \textbf{Model} & \textbf{Accuracy} & \textbf{AUC} & \textbf{F1} \\
\midrule
Video & GenConViT \cite{wodajo2023genconvit}& 0.75 & 0.82 & 0.71 \\
      & FTCN \cite{zheng2021ftcn} & 0.65 & 0.71 & 0.62 \\
      & Styleflow \cite{choi2024styleflow}& 0.53 & 0.56 & 0.58 \\
\midrule
Audio & AASIST \cite{Jung2021AASIST}& 0.84 & 0.91 & 0.78 \\
      & RawNet2 \cite{tak2021endtoendantispoofingrawnet2}& 0.82 & 0.88 & 0.86 \\
      & P3 \cite{wang2023largescalevocodedspoofeddata_model-id-p3}& 0.86 & 0.92 & 0.81 \\
\midrule
Image & UFD \cite{ojha2023fakedetect_ufd}& 0.63 & 0.56 & 0.77 \\
      & DistilDIRE \cite{lim2024distildire}& 0.61 & 0.56 & 0.73 \\
      & NPR \cite{tan2023rethinking_npr}& 0.69 & 0.73 & 0.76 \\
\bottomrule
\end{tabular}
}

%% file: tables/commercial-results.tex
\caption{Best Commercial Model Performance on Deepfake-Eval-2024}
\centering
\begin{tabular}{lccccc}
\toprule 
\textbf{Modality} & \textbf{Accuracy} & \textbf{AUC} & \textbf{Precision} & \textbf{Recall} & \textbf{F1} \\
\midrule 
Video & 0.78 & 0.79 & 0.77 & 0.77 & 0.77 \\
Audio & 0.89 & 0.93 & 0.89 & 0.84 & 0.87 \\
Image & 0.82 & 0.90 & 0.99 & 0.71 & 0.83 \\
\bottomrule 
\end{tabular}

%% file: sections_cvpr/5_errors.tex
\section{Error Analysis}\label{sec:error-analysis}

\textbf{Error analysis methodology}. To further investigate detection model failures, we identify media traits associated with errors. We perform manual error analysis on the entire finetuning test set of videos and images. Due to the length of the Deepfake-Eval-2024 audio test set, we were unable to evaluate its entirety and instead we manually evaluate a class-balanced random sample of 10\%, consisting of 2000 four-second audio clips. For each modality we identify media traits that are associated with a statistically significant decrease in accuracy as measured by chi-squared tests for each model separately, using significance threshold $p < 0.05$. We exclude models that predict all data as belonging to a single class from this error analysis (off-the-shelf and finetuned UFD \cite{ojha2023fakedetect_ufd} and off-the-shelf P3~\cite{wang2023largescalevocodedspoofeddata_model-id-p3}).

\textbf{Off-the-shelf models perform worse on diffusion-generated videos}. Off-the-shelf GenConViT and FTCN have an average 21.3\% lower accuracy on videos which appear to be generated by a diffusion model (\eg, OpenAI's Sora). (Videos were identified as likely diffusion-generated through the presence of watermarks and diffusion-associated visual characteristics \cite{kamali2024how}.) After finetuning on Deepfake-Eval-2024, which includes other diffusion videos, the accuracy gap narrows to 5.4\%, suggesting that the off-the-shelf models' underperformance was primarily due to a domain shift.

\textbf{Models are challenged by videos with selective facial manipulation and videos with non-facial manipulations}. Deepfake-Eval-2024 includes atypical manipulation patterns such as videos with a mix of real and fake faces, and manipulations in non-facial regions. This differs from conventional datasets that contain either entirely real or entirely fake content. Videos with selective face manipulation where some faces are AI-manipulated while others remain real show a 31\% decrease in accuracy. Videos with non-facial manipulation (\eg, altered objects or locations, body modifications) experience a 17.4\% decrease in accuracy compared to other videos. These performance drops likely stem from the fundamental design of most video detection models, which typically assume that the AI manipulation exists for the entire video, and that the fake areas are confined to faces. Consequently, even after finetuning, these types of manipulations remain particularly challenging for models to detect, with accuracy deficits of 16.85\% for selective face manipulation and 35.47\% for non-facial manipulation as compared to the complementary groups.

\textbf{Models perform worse on audio with non-English languages, silences, and background noise, specifically music}. We focus exclusively on the errors of finetuned audio models because floor effects associated with low performance (Supp. Table \ref{append-tab:off-shelf-combined}) make off-the-shelf performance indistinguishably poor across traits. In finetuned models, non-English audio has an average accuracy that was 7.21\% lower than English audio. Because Deepfake-Eval-2024 is an in-the-wild dataset, some parts of the included audio files are silent. Models have 35.39\% worse accuracy on audio clips that are silent. This is expected behavior, as audio without speakers is out of distribution for models targeting deepfake audio. We also note that models perform worse on audio with background noises (accuracy decrease of 7.66\%). Music in the background is associated with a drop in accuracy of  17.94\%, and a large increase of 26.12\% in false negative rate. Adding background music is a common technique in deepfake generation, and our results suggest that current models often fail to identify fakes when the fakes have music added. The inability of models to accurately predict audio with music is a major vulnerability in existing audio deepfake detection models.

\textbf{Models have lower accuracy on images with text overlays}. We identify image categories such as depicted crowds, skin color, and the presence of text overlays. We did not find any statistically significant differences in performance associated with these categories, but we do observe a decrease in accuracy when images have text: accuracy of finetuned models decreases by 9\% and the F1-score decreases by 10.5\% on average. This indicates a distributional mismatch with existing training datasets, which do not include images with text overlays. We hypothesize that the lack of statistically significant differences in the image models is due to floor effects, as the average accuracy of off-the-shelf and finetuned audio models was low (Supp. Tables \ref{append-tab:off-shelf-combined}, \ref{append-tab:finetune-combined}).

\textbf{Many errors are not attributable to human-identifiable characteristics}. The identified error-associated media characteristics do not encompass all errors in the dataset (\eg, an average of 33\% of audio errors have music in the background, and an average 5\% of the image errors had text overlays). The errors not associated with media traits are caused by other failures in model signal interpretation. Developing methods to identify non-visible error patterns is an important area of future research. 

%% file: sections_cvpr/6_conclusion.tex
\section{Discussion \& Limitations}\label{sec:discussion}

In-the-wild deepfake data is crucial for evaluating detection models against real-world threats. However, we acknowledge that curating and manually labeling in-the-wild data is costly and susceptible to human error, resulting in a dataset size appropriate for evaluation but insufficient for wide-scale training. As such, there is still a need for synthetic deepfake datasets, and we recommend that these should strive to be more representative of real-world data and contain the media characteristics that our study reveals to be associated with model errors.

While Deepfake-Eval-2024 is the most comprehensive and diverse collection of real-world deepfakes available today, the rapid evolution of generative AI means that datasets can quickly become outdated, and more work is needed to develop systems that track emerging deepfakes and regularly update datasets. There is also a risk of adversarial actors using Deepfake-Eval-2024 to develop new deepfake generation techniques that evade detectors. To address this risk we make the dataset available at \TrueMedia.org with a CC-BY-SA-4.0 license, but gate access to individuals who are verifiably working on deepfake detection or at research institutions (See Appendix \ref{app:dataset_access}). %

Ultimately, we believe the release of Deepfake-Eval-2024 as a benchmark for contemporary real-world deepfake detection will catalyze the development of robust models that can effectively address the evolving threat of modern deepfakes.

%% file: sections_cvpr/8_acknowledgements.tex
\section{Acknowledgements}
This work was made possible through the incredible team at TrueMedia.org, including Alex
Schokking, Art Min, Dawn Wright, Field Cady, James Allard, Kathy Thrailkill, Maryvel Dolotanora,
Maxwell Bennett, Michael Bayne, Michael Langan, Molly Norris Walker, Paul Carduner, and Steve
Geluso. We would like to thank Ranjay Krishna and Ludwig Schmidt for their advice and Camp.org
for the generous funding that supported this work.

%% file: sections_cvpr/9_supp_materials.tex
\clearpage
\appendix
\setcounter{page}{1}
\maketitlesupplementary
\onecolumn
\renewcommand{\thetable}{S\arabic{table}}
\renewcommand{\thefigure}{S\arabic{figure}}
\setcounter{table}{0}
\setcounter{figure}{0}

\makeatletter
\renewcommand{\theHtable}{appendix.\arabic{table}}
\makeatother
\makeatletter
\renewcommand{\theHfigure}{appendix.\arabic{figure}}
\makeatother

\section{Appendix: Supplementary Materials}

\subsection{Related Work Supplementary Figures}\label{app:related-work-section}
Here we provide a detailed overview of popular deepfake detection datasets and compare them to Deepfake-Eval-2024. We focus on popular datasets released prior to the release of Deepfake-Eval-2024 (March 4, 2025).   

\subsubsection{Overlap between Modality Datasets}\label{app:overlap}
Most video datasets only include manipulated or AI-generated frames from videos without accompanying real or fake audio \cite{faceforensicspp, celeb_df, zi2020wilddeepfake, jiang2020deeperforensics1}, while a few datasets provide audio-visual (AV) data \cite{khalid2021fakeavceleb, cai2024av-deepfake1m, barrington2024deepspeakdatasetv10}. For datasets with AV data, if it is possible to separate audio and video components and labels, we denote the datasets in Tables \ref{tab:related-work_video} and \ref{tab:related-work_audio} with (A) or (V) to describe which part of the datasets we are reporting on. Similarly, there is often overlap between video and image datasets; some popular datasets used for image deepfake detection training and evaluation are composed of individual frames from video datasets \cite{faceforensicspp, zi2020wilddeepfake}. To avoid reporting duplicate datasets across modalities, we omit these from Table \ref{tab:related-work_image}. 

\begin{table*}[h!]
    \input{tables/related-work_video}

\end{table*}
\begin{table*}[ht]
    \input{tables/related-work_audio}

\end{table*}
\begin{table*}
    \input{tables/related-work_image}

\end{table*}

\FloatBarrier
\subsection{Dataset Supplementary Figures}
\begin{table*}[h!]
   \caption{Deepfake-Eval-2024 Video Summary Statistics}
  \label{append-tab:video_stats}
  \makebox[\linewidth]{
      \centering
      \begin{tabular}{lccccc}
        \toprule
        \textbf{Category} & \textbf{Total Duration (hrs)} & \textbf{Count}   & \textbf{Avg. Duration (s)} & \textbf{Avg. FPS} & \textbf{Mode Resolution (W$\times$H)} \\
        \midrule
        Real   & 28.9 & 1,072 & 96.94 & 30.92 & 1,280$\times$720 \\
        Fake   & 16.2 & 964 & 60.47 & 29.09 & 576$\times$720 \\
        All    & 45.1 & 2,036 & 79.68 & 30.05 & 576$\times$720 \\
        \bottomrule
      \end{tabular}
  }
\end{table*}

\begin{table*}
  \caption{Deepfake-Eval-2024 Audio Summary Statistics}
  \label{append-tab:audio_stats}
  \centering
  \begin{tabular}{lcccc}
    \toprule
    \textbf{Category} & \textbf{Total Duration (hrs)}  & \textbf{Count} &  \textbf{Avg. Duration (s)} & \textbf{Avg. Sampling Rate (kHz)} \\
    \midrule
    Real   & 36.6 & 1,110 & 124.80 & 44.83 \\
    Fake  & 19.9 &  710 & 101.51 & 44.40 \\
    All    & 56.5 & 1,820 & 115.46 & 44.66 \\
    \bottomrule
  \end{tabular}
\end{table*}

\begin{table}
  \caption{Deepfake-Eval-2024 Image Summary Statistics}
  \label{append-tab:image_stats}
  \centering
  \begin{tabular}{lcc}
    \toprule
    \textbf{Category} & \textbf{Count} & \textbf{Mode Resolution} \\
    \midrule
    Fake  & 1,208 & 1,200\ensuremath{\times}1,200 \\
    Real  &  767 & 1,024\ensuremath{\times}1,024 \\
    All   & 1,975 & 1,024\ensuremath{\times}1,024 \\
    \bottomrule
  \end{tabular}
\end{table}

\FloatBarrier

\begin{figure*}[h]
    \centering
    \begin{minipage}{0.3\textwidth}
        \centering
        \includegraphics[width=\textwidth]{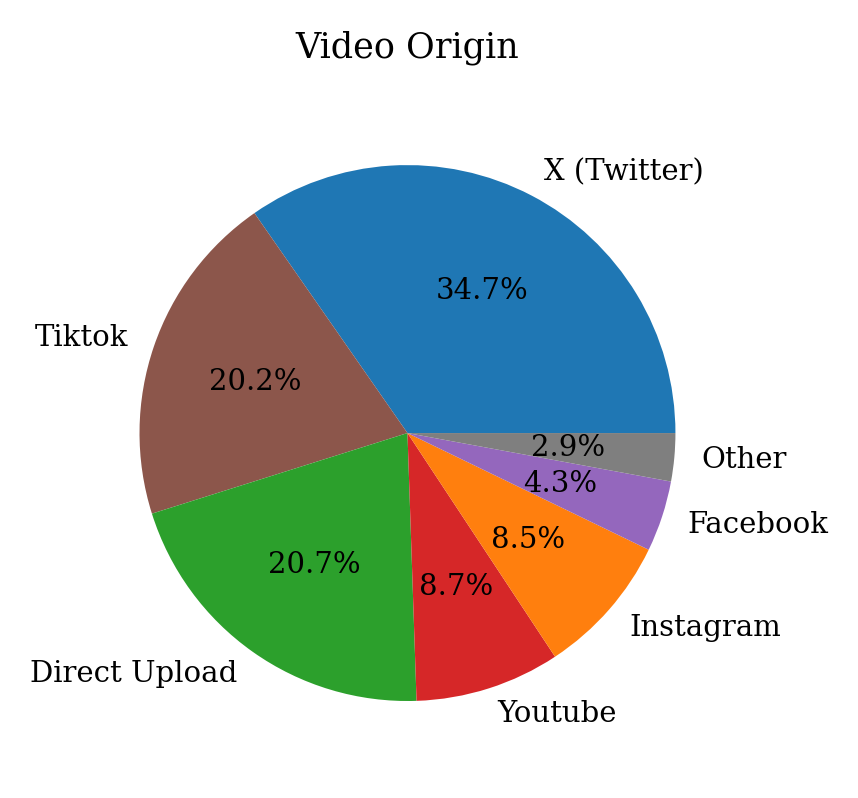}
    \end{minipage}
    \hfill
    \begin{minipage}{0.32\textwidth}
        \centering
        \includegraphics[width=\textwidth]{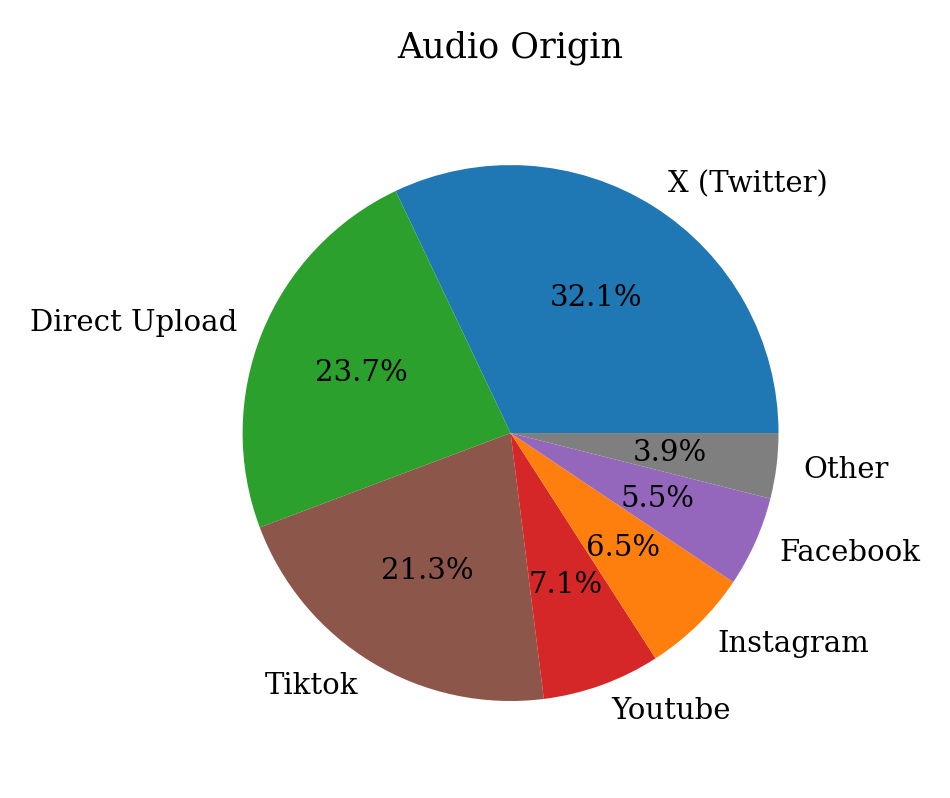}
    \end{minipage}
    \hfill
    \begin{minipage}{0.29\textwidth}
        \centering
        \includegraphics[width=\textwidth]{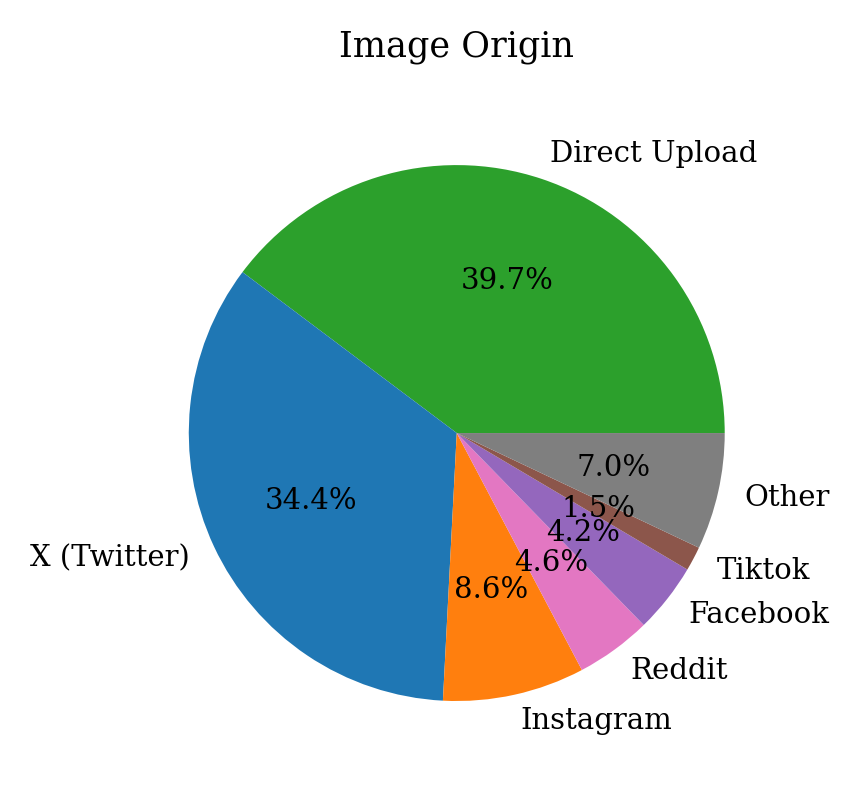}
    \end{minipage}
    \caption{Origins of data in Deepfake-Eval-2024 separated by modality. In total, media was shared from 88 different web-domain names. Direct upload indicates that the media was uploaded directly to \TrueMedia.org by a user, instead of the user providing a link to a social media website.}
    \label{ext-fig:all-origins-separate}
\end{figure*}

\begin{figure*}[h!]
    \centering
    \begin{minipage}{0.45\textwidth}
        \centering
        \includegraphics[width=\textwidth]{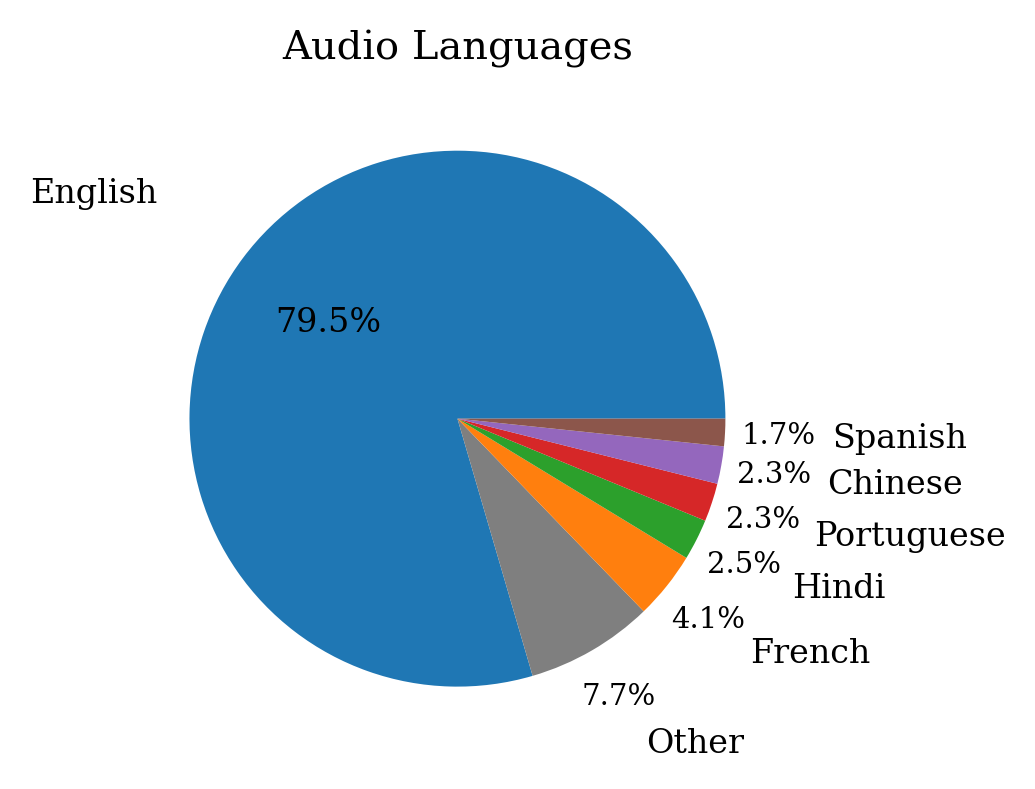}
    \end{minipage}
    \hfill
    \begin{minipage}{0.45\textwidth}
        \centering
        \includegraphics[width=\textwidth]{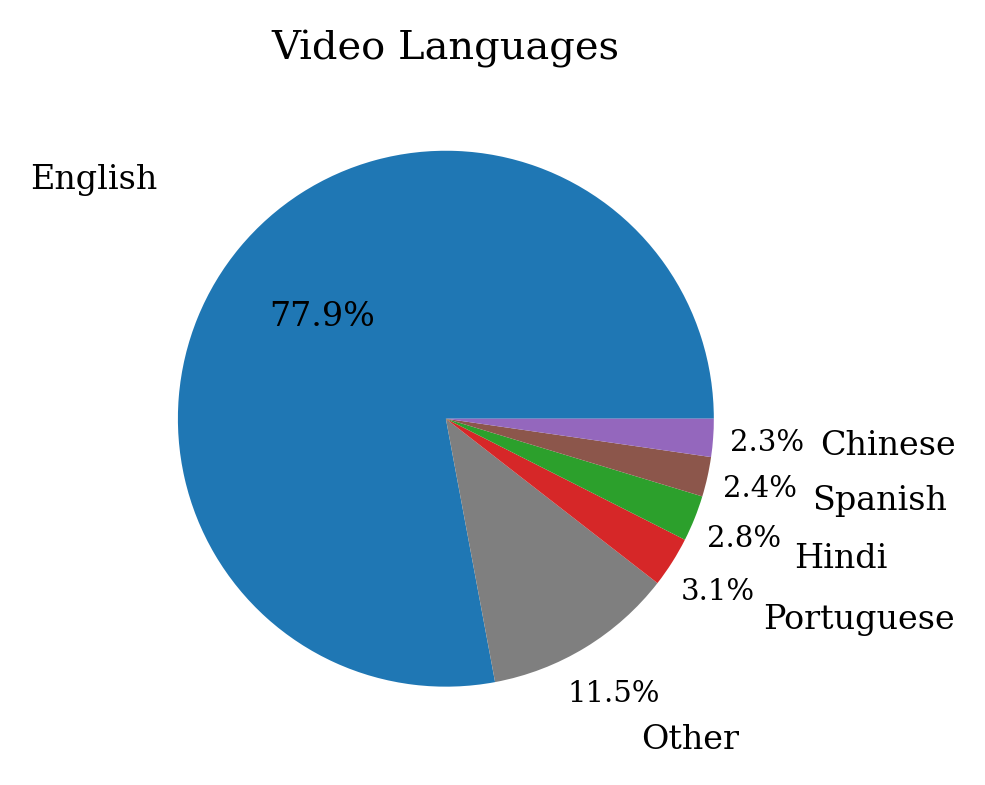}
    \end{minipage}
    \caption{Language distributions for audio and video content.}
    \label{ext-fig:language-distributions-separated}
\end{figure*}

\subsubsection{Demographic Estimates of Image and Video Subjects}\label{app:demographic-estimates}

\begin{table*}
    \input{tables/rebuttal-demographic-breakdown}
\end{table*}

To characterize the demographic composition of Deepfake-Eval-2024, we manually annotated a randomly sampled 10\% subset of the visual data for perceived race/ethnicity. We report the demographic estimates from this analysis in Supp. Table \ref{append-tab:demographic-estimates}. We do not make definitive claims about subject identity. 

\subsection{Experimental Compute}\label{app:compute}
We evaluated and finetuned all models on a single AWS A10 GPU on an g5.2xlarge instance or a single GCP L4 GPU on a g2-standard-8 instance. We found no difference between evaluating and finetuning models on the AWS or GCP instance.

\subsection{Results Supplementary Figures}
\FloatBarrier
\begin{table*}[h]
    \input{tables/appendix-model-results-combined-DFE24}
    \label{append-tab:off-shelf-combined}
    \vspace{1em}  %
    \input{tables/appendix-finetune-result-combined}
    \label{append-tab:finetune-combined}
\end{table*}

\begin{table*}
\input{tables/rebuttal_commercial_table}
\end{table*}

\begin{table*}
    \input{tables/rebuttal-multimodal-results}
\end{table*}

\subsection{Dataset Access and Ethics}\label{app:dataset_access}
 Deepfakes pose an established threat to society, which is why the release of updated deepfake detection benchmarks is important for improving our response to this threat. However, there is a potential for released datasets to be used with malicious intentions to create deepfake generation technologies that are more realistic and that evade existing detectors. As such, we gate access to this dataset through Huggingface to individuals verifiably at research institutions or doing work related to deepfake detection. Prospective users must accept a terms of use agreement. Then they must provide an an institutional / company email address and link that provides additional evidence of work related to deepfake detection, which is used for verification. We intend to maintain this system of access gating until May 2027, at which point generative AI is likely to have advanced far beyond the deepfakes represented in Deepfake-Eval-2024, and thus there will be a much lower risk of making Deepfake-Eval-2024 available to all individuals. Starting in May 2027, we will modify the access process so that any individual who agrees to the Terms of Use can get access to the dataset. Further, in the Terms of Use, we state that we release the dataset with a CC-BY-SA-4.0 license, designating it primarily as a research artifact. While the dataset is freely available, users are responsible for ensuring its legal use in commercial settings. Users must independently verify compliance with applicable laws before employing the dataset for commercial purposes.

\onecolumn
\section{Appendix: Labeling Criteria}\label{app:labeling-criteria}
We present the labeling criteria for all modalities. The complementary examples mentioned in this section can be found in Supplementary .zip file. 

\subsection{Image labeling codebook}
AI-generated video/image traits adapted from Kamali et al. \cite{kamali2024how}.

\begin{longtable}{|p{0.3\textwidth}|p{0.3\textwidth}|p{0.3\textwidth}|}
\hline
\multicolumn{1}{|c|}{\textbf{Real (no AI manipulation)}} & 
\multicolumn{1}{c|}{\textbf{Fake (AI manipulation)}} & 
\multicolumn{1}{c|}{\textbf{Unknown}} \\ \hline

Original, reputable source confirms no AI manipulation & If any portion is AI, then entire item is fake & Cartoons, animations, and photoshopped images such as swapped signs, hats, or t-shirts (unless evidence of AI manipulation) \\ \hline

Fact-checking source confirms no AI manipulation & Fact-checking source confirms AI manipulation & Unable to confirm AI manipulation or not \\ \hline

Real media in which a person is lying, or real images presented out of context and misleading & Contains 3+ of the following AI traits: 
\begin{itemize}[left=0pt]
    \item Stylistic Artifacts: hyper-realistic or inconsistent detail, smooth or plastic/waxy looking skin (\href{}{Example 1}), cartoonish appearance (\href{}{Example 2}), too perfect, inconsistent lighting or reflections etc.
    \item Anatomical Implausibilities: irregular pupils, mangled/ missing/disproportionate limbs, incorrect/merged fingers, inconsistent facial features of famous personas compared to their real images etc.
    \item Sociocultural Implausibilities: unlikely scenarios or historical inaccuracies
    \item Functional Implausibilities: misspelled/backwards text, impossible words, impossible structure of buildings, vehicles, food etc.
\end{itemize} & \\ \hline

 & Face swapping and face morphing for media created in 2023 or later & Face swapping and face morphing for media created prior to 2023  \\ \hline

Content from film or TV with no evidence of AI manipulation &  &  \\ \hline

Media manipulation using text and non-AI-generated image overlays such as stickers (\href{}{Example 3} and \href{}{Example 4})  & & \\ \hline

\end{longtable}
\newpage
\subsection{Video Codebook}

AI generated video/image traits adapted from Kamali et al. \cite{kamali2024how}

\begin{longtable}{|p{0.3\textwidth}|p{0.3\textwidth}|p{0.3\textwidth}|}
\hline
\multicolumn{1}{|c|}{\textbf{Real (no AI manipulation)}} & 
\multicolumn{1}{c|}{\textbf{Fake (AI manipulation)}} & 
\multicolumn{1}{c|}{\textbf{Unknown}} \\ \hline

Lips and mouth are crisp, clear, nuanced, and match sound perfectly. & Lips are roughly in sync \href{}{Example 5} with audio, but clearly not crisp or natural & \\ \hline

Lips and audio are completely out of sync \href{}{Example 6}, (and you find original source to confirm that audio was dubbed onto a real video) &   & Lips and audio are completely out of sync, but you cannot find the original source to confirm if video is real or manipulated \\ \hline

Located original source and confirmed no AI manipulation & Located original source and confirmed AI manipulation was used  & Video quality is too poor to determine if mouth movements are crisp and nuanced \\ \hline

Highly edited \href{}{Example 7}, but every individual clip is real &   & Filters \href{}{Example 8}, effects, GIFs  \\ \hline

Real person is obviously “lip syncing” \href{}{Example 9} or parody, no evidence of AI manipulation. &  &  \\ \hline

Talking head \href{}{Example 10} pasted on background (predominant in many tiktok videos) & & \\ \hline

\end{longtable}

\subsection{Audio Codebook}

\begin{longtable}{|p{0.3\textwidth}|p{0.3\textwidth}|p{0.3\textwidth}|}
\hline
\multicolumn{1}{|c|}{\textbf{Real (no AI manipulation)}} & 
\multicolumn{1}{c|}{\textbf{Fake (AI manipulation)}} & 
\multicolumn{1}{c|}{\textbf{Unknown}} \\ \hline

Lips and mouth are crisp, clear, nuanced, and match sound perfectly. & If lip sync is off AND 2 or more audio models say $>80$\% & If lip sync is off and you cannot discern if AI or human impersonator \\ \hline

Audio without speech such as music, silence, and sound effects were labeled as real unless there was other evidence of AI manipulation. & Audio-Only: if 2 or more models say $>80$\% likelihood of fake PLUS there’s some additional reason to believe it’s fake (ie. the audio quality sounds synthetic, or sociocultural implausability) \href{}{Example 11}  & Voice is off camera and unable to locate original source \\ \hline

Human impersonator \href{}{Example 12} &   &   \\ \hline

\end{longtable}
\clearpage
\onecolumn

\section{Appendix: Verification Process} \label{app:verification-process}

Reverse Image Search
\begin{itemize}
    \item If a media item did not contain common AI traits to help us determine ground truth, we used Google’s reverse image search to locate the original source of the item, or to find a professional fact-checking source that confirmed the item’s ground truth.
\end{itemize}

Source Trustworthiness
\begin{itemize}
    \item When we located the original source or fact-checking source for an item, we used tools such as All Sides and Ad Fontes Media Bias to judge the trustworthiness of the source before determining the ground truth.
\end{itemize}

ChatGPT
\begin{itemize}
    \item While we did not trust GPT implicitly, we did use it to point us in the right direction. For
example, if a video showed Kamala Harris saying ``xyz,'' we used the following prompt as a first step to determine its veracity: ``Did Kamala Harris say `xyz?' Give me 3 reputable
sources confirming or denying this claim.''
\end{itemize}

Google
\begin{itemize}
    \item We used Google Search to find primary sources confirming or denying media claims. For example, if a video showed Donald Trump saying ``they’re eating the pets of the people who live there,'' we ran the search ``Did Trump say \ldots'' Or, if an image or video depicted Joe Biden falling asleep at a press conference, we ran the search ``Did Biden fall asleep at \ldots'' The results often pointed us to primary sources that we used to determine ground truth.
\end{itemize}

\subsection{Reverse Image Search Verification Process}

\begin{table}[h!]
    \centering
    \begin{tabular}{|p{0.3\textwidth}|p{0.3\textwidth}|p{0.3\textwidth}|}
         \toprule
         & Click on ``See Exact Matches'' & \\
         \midrule
         No Match Found & If no match and no clues, mark as ``Unknown'' & \\ \midrule
         Match on Unknown Source & If match is found on a lesser-known site, check if the image is credited to a reputable source (AP, Reuters, etc.) & If credited, confirm by checking the site. If the site is legitimate, mark as ``Real.'' \\ \midrule
         Match on Social Media & If found on social media, read comments for clues. & If comments suggest it is fake due to artifacts in the media, mark as ``Fake.'' If credible, mark as ``Real.'' \\ \midrule
         Verified Source & If found on a reputable source's social media (NBC, White House, etc.), mark as ``Real.'' & \\ \midrule
         Edited Media & If you find edited media (\eg, face swapped or text altered in a sign), pay close attention to details. & Mark as ``Fake.'' \\ \bottomrule
    \end{tabular}
    \label{tab:appendix-reverse-search}
\end{table}

%% file: tables/related-work_video.tex
  \caption{Survey of existing popular video deepfake detection datasets.}
  \label{tab:related-work_video}
  \makebox[\linewidth]{
  \centering
  \begin{tabular}{m{3cm}cz{1.2cm}z{1.2cm}z{1.5cm}z{1.5cm}z{1.5cm}z{1cm}}
    \toprule
        \textbf{Dataset} 
        & \textbf{Year} 
        & \textbf{\# Real Files}   
        & \textbf{\# Fake Files} 
        & \textbf{Real Media Duration (hrs) }
        & \textbf{Fake Media Duration (hrs) }
        & \textbf{Total Duration (hrs) }
        & \textbf{In-the-Wild} \\
    \midrule
    FaceForensics++ \cite{faceforensicspp}           & 2019 & 1,000   & 4,000   & 4.71\textsuperscript{*}  & 16.95\textsuperscript{*}  & 21.66\textsuperscript{*}       & \textcolor{red}{\ding{55}}  \\
    Celeb-DF \cite{celeb_df}           & 2019 & 590 & 5,639                        & 2.13\textsuperscript{$\dagger$} & 20.36\textsuperscript{$\dagger$} & 22.49\textsuperscript{$\dagger$} & \textcolor{red}{\ding{55}}                   \\
    DFDC \cite{DFDC2020}                   & 2020 & 23,654  & 104,500 & 64.43  & 288.88  & 353.31       & \textcolor{red}{\ding{55}}  \\
    WildDeepfake  \cite{zi2020wilddeepfake}              & 2020 & 3,805   & 3,509   & -      & -       & 10.93\textsuperscript{*}       & \textcolor{green}{\ding{51}} \\
    DeeperForensics-1.0 \cite{jiang2020deeperforensics1}        & 2020 & 50,000  & 10,000  & 46.30\textsuperscript{*} & 116.67\textsuperscript{*} & 162.96\textsuperscript{*}     & \textcolor{red}{\ding{55}}  \\
    DF-W    \cite{pu2021deepfake_dfw}                    & 2021 & 0      & 1,869   & 0      & 48.83   & 48.83        & \textcolor{green}{\ding{51}}  \\
    ForgeryNet \cite{he2021forgerynet}                  & 2021 & 99,630  & 121,617 & 13.32\textsuperscript{*} & 13.50\textsuperscript{*}  & 26.82\textsuperscript{*}       & \textcolor{red}{\ding{55}}  \\
    FakeAVCeleb (V) \cite{khalid2021fakeavceleb} & 2021 & 500 & 19,000                       & 1.08\textsuperscript{$\dagger$} & 41.17\textsuperscript{$\dagger$} & 42.25\textsuperscript{$\dagger$} & \textcolor{red}{\ding{55}}                     \\
    GOTCHA \cite{mittal2023gotcha}                     & 2022 & 409    & 55,838  & 3.13\textsuperscript{$\ddagger$} & -       & -            & \textcolor{red}{\ding{55}}  \\
    RWDF-23 \cite{cho2023towards}           & 2023 & 0    & 2,000 & -      & 48.15       & 48.15 & \textcolor{green}{\ding{51}}  \\
    DF-Platter \cite{df_platter}                 & 2023 & 764    & 132,496 & -      & -       & $\approx$736.08 & \textcolor{red}{\ding{55}}  \\
    AV-Deepfake1M \cite{cai2024av-deepfake1m} & 2023 & 286,721 & 860,039 & -      & -       & 1,886         & \textcolor{red}{\ding{55}}  \\
    DeepSpeak \cite{barrington2024deepspeakdatasetv10} & 2024 & 6,226   & 6,799   & 17     & 26      & 44           & \textcolor{red}{\ding{55}}  \\
    DF40 \cite{yan2024df40}                       & 2024 & 0      & 100k+ & -      & -       & -            & \textcolor{red}{\ding{55}} \\
    \midrule
    \textbf{Ours}               & 2024 & 1,072   & 964    & 28.9  & 16.2      & 45.1         &  \textcolor{green}{\ding{51}} \\
    \bottomrule
  \end{tabular}
  }
  \tabletext{When duration values are not directly provided, values are estimated using several methods: \textsuperscript{*} indicates calculation from frame count assuming 30fps (the most commonly encountered frame rate among published video datasets), \textsuperscript{$\dagger$} indicates derivation from average clip lengths, \textsuperscript{$\ddagger$} indicates values estimated from reported estimates, and $\approx$ indicates direct reported estimates.}

%% file: tables/related-work_audio.tex
  \caption{Survey of existing popular audio deepfake detection datasets.}
  \label{tab:related-work_audio}
  \makebox[\linewidth]{
  \centering
  \begin{tabular}{p{2.5cm}cz{1.8cm}z{2cm}z{1.2cm}z{1.2cm}z{1.2cm}z{1cm}z{1cm}}
    \toprule
        \textbf{Dataset }
        & \textbf{Year} 
        & \textbf{\# Real Files}   
        & \textbf{\# Fake Files }
        & \textbf{Real Media (hrs) }
        & \textbf{Fake Media (hrs) }
        & \textbf{Total Duration (hrs) }
        & \textbf{In-the-Wild }
        & \textbf{\# Languages} \\
    \midrule
    FoR \cite{Reimao2019FoRAD}                      & 2019 & 108,256        & 87,285         & 151.86\textsuperscript{$\dagger$}     & 56.98\textsuperscript{$\dagger$}      & 208.84\textsuperscript{$\dagger$}                  & \textcolor{red}{\ding{55}}          & 1                                                      \\
    ASVspoof (LA subset) \cite{wang2020asvspoof2019, yamagishi2019asvspoof_evalplan} & 2019 &  12,483 & 108,978 & 5.20\textsuperscript{$\dagger$} & 45.41\textsuperscript{$\dagger$} & 50.61\textsuperscript{$\dagger$} & \textcolor{red}{\ding{55}}  & 1 \\
    FakeAVCeleb (audio) \cite{khalid2021fakeavceleb}      & 2021 & 500           & 10,500         & 1.08\textsuperscript{$\dagger$}      & 22.75\textsuperscript{$\dagger$}     & 23.83\textsuperscript{$\dagger$} & \textcolor{red}{\ding{55}}        & 1                                                    \\
    WaveFake \cite{frank2021wavefake}                 & 2021 & 0             & 117,985        & 0                            & $\approx$196                    & $\approx$196               & \textcolor{red}{\ding{55}}          & 2                                           \\
    ASVspoof (DF subset) \cite{Liu_2023_asvspoof21} & 2021 & 20,637         & 572,616        & -                            & -                            & 325.8\textsuperscript{\S}                 & \textcolor{red}{\ding{55}}         & 1                                                    \\
    In-the-Wild \cite{muller2022does_in-the-wild}              & 2022 & -             & -             & 20.7                         & 17.2                         & 37.9                    & \textcolor{green}{\ding{51}}         & 1                                                    \\
    SpoofCeleb~\cite{jung2024spoofceleb}             & 2024 & $\approx$ 248,000           & $\approx$ 2,439,292            & 310\textsuperscript{$\dagger$}                         & 3,049\textsuperscript{$\dagger$}                         & 3,359\textsuperscript{$\dagger$}                    & \textcolor{red}{\ding{55}}          & 1                                                    \\
    \midrule 
    \textbf{Ours }                     & 2024 & 1,167          & 814           & 36.6                         & 19.9                         & 56.5                    & \textcolor{green}{\ding{51}}         & 42 \\

    \bottomrule
  \end{tabular}
  }
  \tabletext{Datasets that are not publicly available yet (such as ASVspoof5) are not included. Similar to video datasets, when duration values are not directly provided, values are estimated using several methods: \textsuperscript{$\dagger$} indicates derivation from average clip lengths, $\approx$ indicates direct reported estimates, and \textsuperscript{\S} indicates values provided by a survey paper \cite{yi2023audiodeepfakedetectionsurvey}.}

%% file: tables/related-work_image.tex
  \caption{Survey of existing popular image deepfake detection datasets.}
  \label{tab:related-work_image}
  \makebox[\linewidth]{
  \centering
  \begin{tabular}{m{2.5cm}cz{1.2cm}z{1.2cm}z{1.2cm}z{1.2cm}z{2cm}c}
    \toprule
        \textbf{Dataset }
        & \textbf{Year }
        & \textbf{\# Real Files}   
        & \textbf{\# Fake Files} 
        & \textbf{\# Total Files}
        & \textbf{In-the-Wild}
        & \textbf{\# Generation Techniques}
        & \textbf{Resolution} \\
    \midrule
    iFakeFaceDB \cite{9133490_ifake}        & 2019 & 0       & $\approx$87,000 & $\approx$87,000 & \textcolor{red}{\ding{55}}  & 2    & {224\ensuremath{\times}224}   \\
    DFFD \cite{cvpr2020-dang_dffd}              & 2020 & 58,703   & 240,336      & 299,039      & \textcolor{red}{\ding{55}}  & 4    & {1,024\ensuremath{\times}1,024} \\
    ForenSynths \cite{wang2019cnngenerated_forensynth}       & 2020 & 36,200   & 36,200       & 72,400       & \textcolor{red}{\ding{55}}  & 11   & {256\ensuremath{\times}256}  \\
    ForgeryNet (image) \cite{he2021forgerynet} & 2021 & 1,438,201 & 1,457,861     &  2,896,062       & \textcolor{red}{\ding{55}}  & 15   & Varies    \\
    DiffusionForensics \cite{wang2023dire_diffusionforensics} & 2023 & 232,000  & 232,000      & 464,000      & \textcolor{red}{\ding{55}}  & 11   & {256\ensuremath{\times}256}   \\
    CIFAKE \cite{Bird2023CIFAKEIC}            & 2024 & 60,000   & 60,000       & 120,000      & \textcolor{red}{\ding{55}}  & 1    & {32\ensuremath{\times}32}     \\
    \midrule
    \textbf{Ours}      & 2024 & 767     & 1,208        & 1,975        & \textcolor{green}{\ding{51}} & Many & Varies \\
    \bottomrule
  \end{tabular}
  }

%% file: tables/rebuttal-demographic-breakdown.tex
\caption{Demographic estimates of image and video subjects in Deepfake-Eval-2024.}
\label{append-tab:demographic-estimates}
\centering
\begin{tabular}{lc}
\toprule 
\textbf{Race/Ethnicity} & \textbf{Percentage} \\
\midrule
White & 54.6 \\
Multiple & 16.8 \\
Black & 9.2 \\
East Asian & 6.6 \\
South Asian & 5.3 \\
Unknown - non-white & 5.0 \\
Latino - non-white & 2.4 \\
\bottomrule 
\end{tabular}

%% file: tables/appendix-model-results-combined-DFE24.tex
\caption{Complete Off-the-Shelf Open-Source Model  Results Across Modalities}
\centering
\begin{tabular}{llcccccccc}
\toprule
\textbf{Modality} & \textbf{Model} & \textbf{AUC} & \textbf{Accuracy} & \textbf{Precision} & \textbf{Recall} & \textbf{F1} & \textbf{FPR} & \textbf{FNR} & \textbf{EER (\%)} \\
\midrule
Video & GenConViT & 0.63 & 0.60 & 0.60 & 0.50 & 0.54 & 0.31 & 0.50 & - \\
      & FTCN & 0.50 & 0.51 & 0.51 & 0.67 & 0.41 & 0.33 & 0.66 & - \\
      & Styleflow & 0.51 & 0.52 & 0.54 & 0.43 & 0.48 & 0.39 & 0.56 & - \\
\midrule
Audio & AASIST & 0.43 & 0.42 & 0.31 & 0.51 & 0.39 & 0.63 & 0.49 & 55.22 \\
      & RawNet2 & 0.53 & 0.48 & 0.66 & 0.39 & 0.49 & 0.36 & 0.61 & 48.20 \\
      & P3 & 0.58 & 0.36 & 0.36 & 1.00 & 0.53 & 1.00 & 0.00 & 43.00\\
\midrule
Image & UFD & 0.56 & 0.63 & 0.63 & 0.999 & 0.77 & 0.99 & 0.001 & - \\
      & DistilDIRE & 0.52 & 0.61 & 0.64 & 0.87 & 0.74 & 0.83 & 0.13 & - \\
      & NPR & 0.53 & 0.47 & 0.69 & 0.29 & 0.41 & 0.22 & 0.71 & - \\
\bottomrule
\end{tabular}

%% file: tables/appendix-finetune-result-combined.tex
\caption{Complete Open-Source Model Finetuning Results Across Modalities}
\centering
\begin{tabular}{llcccccccc}
\toprule
\textbf{Modality} & \textbf{Model} & \textbf{AUC} & \textbf{Accuracy} & \textbf{Precision} & \textbf{Recall} & \textbf{F1} & \textbf{FPR} & \textbf{FNR} & \textbf{EER (\%)} \\
\midrule
Video & Genconvit & 0.82 & 0.75 & 0.78 & 0.65 & 0.71 & 0.17 & 0.35 & - \\
      & FTCN & 0.71 & 0.65 & 0.64 & 0.61 & 0.62 & 0.30 & 0.39 & - \\
      & Styleflow & 0.56 & 0.53 & 0.52 & 0.66 & 0.58 & 0.61 & 0.34 & - \\
\midrule
Audio & AASIST & 0.906 & 0.836 & 0.797 & 0.761 & 0.778 & 0.118 & 0.239 & 16.99 \\
      & RawNet2 & 0.876 & 0.817 & 0.818 & 0.908 & 0.860 & 0.334 & 0.092 & 20.91 \\
      & P3 & 0.920 & 0.855 & 0.802 & 0.818 & 0.810 & 0.122 & 0.182 & 15.38 \\
\midrule
Image & UFD & 0.56 & 0.63 & 0.63 & 1.00 & 0.77 & 1.00 & 0.00 & - \\
      & DistilDIRE & 0.56 & 0.61 & 0.64 & 0.87 & 0.74 & 0.85 & 0.13 & - \\
      & NPR & 0.55 & 0.58 & 0.61 & 0.81 & 0.70 & 0.76 & 0.19 & - \\
\bottomrule
\end{tabular}

%% file: tables/rebuttal_commercial_table.tex
\caption{Best Commercial Model Performance on Deepfake-Eval-2024}
\label{app-table:commercial-results}
\centering
\begin{tabular}{lcccccc}
\toprule 
\textbf{Modality} & \textbf{Model Rank}& \textbf{Accuracy} & \textbf{AUC} & \textbf{Precision} & \textbf{Recall} & \textbf{F1} \\
\midrule 
Video & \#1 & 0.78 & 0.79 & 0.77 & 0.77 & 0.77 \\
& \#2 & 0.66 & 0.70 & 0.78 & 0.47 & 0.59 \\
&\#3 & 0.59 & 0.64 & 0.59 & 0.69 & 0.64 \\
\midrule
Audio & \#1& 0.89 & 0.93 & 0.89 & 0.84 & 0.87 \\
&\#2 & 0.88 & 0.93 & 0.88 & 0.80 & 0.84 \\
&\#3 & 0.86 & 0.90 & 0.83 & 0.83 & 0.84 \\
\midrule
Image & \#1& 0.86 & 0.88 & 0.97 & 0.79 & 0.87 \\
& \#2& 0.82 & 0.90 & 0.99 & 0.71 & 0.83 \\
& \#3& 0.77 & 0.89 & 0.98 & 0.64 & 0.77 \\
\bottomrule 
\end{tabular}

%% file: tables/rebuttal-multimodal-results.tex
\caption{Open-Source Multimodal Model Results}
\label{append-tab:multimodal-results}
\centering
\begin{tabular}{lcc}
\toprule 
\textbf{Model} & \textbf{AUC on Deepfake-Eval-2024}& \textbf{AUC on Original Publication Test Dataset} \\
\midrule
AVF~\cite{feng2023self} & 0.58 & 0.945 (FakeAVCeleb), 0.87 (KoDF) \\
FGI~\cite{astrid2024detecting} & 0.42 & 0.845 (FakeAVCeleb), 0.98 (DFDC)\\
\bottomrule 
\end{tabular}